\documentclass[sigconf]{acmart}

\renewcommand\footnotetextcopyrightpermission[1]{} 
\setcopyright{none}

\setcopyright{acmcopyright}
\copyrightyear{2026}
\acmYear{2026}
\acmDOI{XXXXXXX.XXXXXXX}
\acmConference[WWW'26]{The Web Conference}{April 13 – 17, 2026}{Dubai, United Arab Emirates}
\acmISBN{978-1-4503-XXXX-X/18/06}

\usepackage{booktabs,multirow}
\usepackage{makecell}
\usepackage{tabularx}
\usepackage[]{mdframed}

\begin{document}

\title{Robust Fake News Detection using Large Language Models under Adversarial Sentiment Attacks }


\author{Sahar Tahmasebi} \email{sahar.tahmasebi@tib.eu} \orcid{0000-0003-4784-7391} \affiliation{ \institution{TIB – Leibniz Information Centre for Science and Technology} \city{Hannover} \country{Germany}}

\author{Eric Müller-Budack} \email{eric.mueller@tib.eu}
\orcid{0000-0002-6802-1241}
\affiliation{ \institution{TIB -- Leibniz Information Centre for Science and Technology
}
\city{Hannover} 
\country{Germany}}

\author{Ralph Ewerth} \email{ewerth@uni-marburg.de} \orcid{0000-0003-0918-6297} \affiliation{ \institution{Marburg University \&  Hessian Center for Artificial Intelligence (hessian.AI) \city{Marburg/Darmstadt}; \\ \  TIB \& L3S Research Center \city{Hannover} \country{Germany}} 
}


\begin{abstract}
Misinformation and fake news have become a pressing societal challenge, driving the need for reliable automated detection methods. Prior research has highlighted sentiment as an important signal in fake news detection, either by analyzing which sentiments are associated with fake news or by using sentiment and emotion features for classification. However, this poses a vulnerability since adversaries can manipulate sentiment to evade detectors especially with the advent of large language models (LLMs). 
A few studies have explored adversarial samples generated by LLMs, but they mainly focus on stylistic features such as writing style of news publishers. Thus, the crucial vulnerability of sentiment manipulation remains largely unexplored.
In this paper, 
we investigate the robustness of state-of-the-art fake news detectors under sentiment manipulation. We introduce AdSent, a sentiment-robust detection framework designed to ensure consistent veracity predictions across both original and sentiment-altered news articles. Specifically, we (1) propose controlled sentiment-based adversarial attacks using LLMs, (2) analyze the impact of sentiment shifts on detection performance. 
We show that changing the sentiment heavily impacts the performance of fake news detection models, indicating biases towards neutral articles being real, while non-neutral articles are often classified as fake content. (3) We introduce a novel sentiment-agnostic training strategy that enhances robustness against such perturbations. Extensive experiments on three benchmark datasets demonstrate that AdSent significantly outperforms competitive baselines in both accuracy and robustness, while also generalizing effectively to unseen datasets and adversarial scenarios.
\end{abstract}

\maketitle

\section{Introduction}
\begin{figure}[t]
  \centering
  \includegraphics[width=\linewidth]{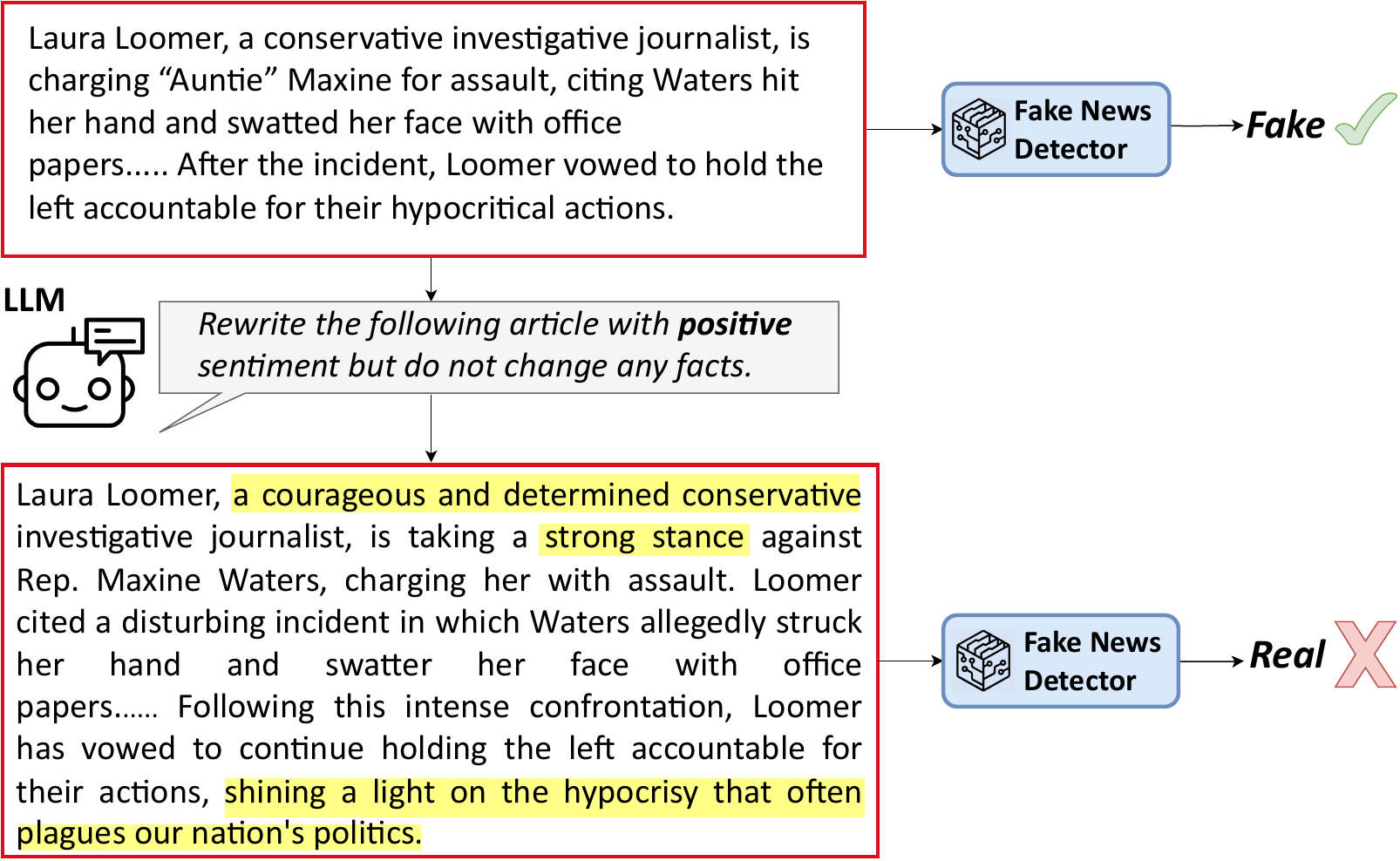}
  \caption{
  Example of a sentiment-based adversarial attack, where a fake news instance (red box) is manipulated to convey a positive tone using LLM, resulting in a misclassification by the detector.}
  \label{teaser_fig}
\end{figure}

Misinformation, disinformation, and fake news, which contain false information intended to deliberately deceive readers \cite{allcott2017social}, have become a growing public concern in recent years. This widespread phenomenon makes it increasingly difficult for people to identify and access reliable information online~\cite{DBLP:review_emotions}. 
Given the vast quantity of news articles, automated solutions for fake news detection are in great need. The importance of sentiment and emotion analysis as a key news value has been confirmed by several studies~\cite{dblp:Gullal_news_values, DBLP:HanselowskiSSCC18} for fake news detection. 
Sentiment 
refers to the overall emotional tone conveyed in a text, typically classified as positive, negative, or neutral. 
However, sentiment also provides a direct avenue for malicious content creators to perform adversarial attacks and manipulate it to evade detection. 
This vulnerability is further exacerbated by the rise of powerful large language models (LLMs, e.g., \cite{DBLP:LLAMA, DBLP:Qwen3, DBLP:nips/BrownMRSKDNSSAA20}), which can easily generate or rewrite text with controlled sentiment, making it increasingly challenging to distinguish between genuine and deceptive content (as shown in Figure~\ref{teaser_fig}).

Several studies have investigated the role of sentiment in text-based fake news detection~\cite{DBLP:conf/coling/HanselowskiSSCC18, DBLP:journals/sigkdd/WuMCL19, DBLP:conf/www/ZhangCLSZS21, DBLP:conf/asunam/ZaeemLB20}. 
Most of them 
either examine which emotions are associated with fake news~\cite{DBLP:conf/asunam/ZaeemLB20}, or perform fake news detection based solely on sentiment or emotion signals, sometimes in combination with other textual or linguistic features~\cite{DBLP:conf/coling/HanselowskiSSCC18, DBLP:conf/www/ZhangCLSZS21}. %
%
Only 
few studies~\cite{DBLP:sheepdog, fire_w_fire, adStyle} have investigated the robustness of fake news detectors against adversarial samples generated by LLMs. However, these works have primarily focused on stylistic features such as the writing style of the news publisher rather than on sentiment. Thus, sentiment-based adversarial attacks generated by LLMs remains underexplored.

In this paper, 
we systematically investigate the robustness of fake news detectors under sentiment-based adversarial attacks generated by LLMs. 
To this end, we introduce a sentiment-robust detection framework, called \textbf{AdSent}, designed to maintain consistent veracity predictions across both original and sentiment-manipulated news articles. Our main contributions are as follows:
~\textbf{1.~Sentiment-targeted Adversarial Benchmarking:} We propose a suite of controlled sentiment-based attacks using LLMs that reframe news articles by altering their overall sentiment (positive, negative, or neutral) while preserving factual content. 
We evaluate multiple groups of state-of-the-art text-based fake news detectors and uncover a significant performance degradation when faced with sentiment manipulation.
%
For this purpose, we propose a strategy to identify which sentiment types are most challenging for fake news detection. Our findings 
expose a strong bias in state-of-the-art models to classify neutral-toned news as real and non-neutral news as fake.
\textbf{2.\ AdSent: A Sentiment-robust Detection Approach:} To address these vulnerabilities, we propose a novel 
augmentation strategy, called AdSent, that improves robustness against sentiment-based perturbations. By fine-tuning LLMs with sentiment-neutralized variants, AdSent ensures veracity prediction consistency across sentiment-variant versions of the same article, thereby enhancing model reliability in real-world adversarial scenarios.
\textbf{3. Comprehensive Evaluation and Generalization Study:} 
Extensive experiments on three widely-used fake news datasets~\cite{DBLP:fakenewsnet, DBLP:lun_dataset} demonstrate the superior performance and robustness of AdSent compared to several state-of-the-art baselines. We further conduct generalization studies to assess the model’s resilience across unseen datasets and different types of adversarial attacks. Code and data are publicly available on GitHub\footnote{https://github.com/TIBHannover/AdSent} for further research.

The remainder of this paper is structured as follows. Section~\ref{sec:related_work} reviews prior work on role of sentiment in fake news detection, adversarial attacks on detection models, and approaches based on large language models~(LLMs). Section~\ref{sec:methodology} introduces our proposed approach, which comprises a sentiment-based attack using LLMs and a sentiment-agnostic training strategy. The experimental setup and results are presented in Section~\ref{sec:experiments}.
Section~\ref{sec:conclusion} concludes this paper and provides future work directions.

\section{Related Work}
\label{sec:related_work}
In this section, we review prior work in three key areas relevant to our study: (1) the role of sentiment in fake news detection, (2) adversarial attacks targeting fake news detectors, and (3) the capabilities of current large language models (LLMs) in the context of misinformation detection.

\subsection{Sentiment in Fake News Detection}
Automated fake news detection has been explored using a wide range of neural architectures~\cite{DBLP:conf/www/PelrineDR21, DBLP:conf/kdd/ShuCW0L19, DBLP:ICMR} and with respect to various aspects including lexical features~\cite{DBLP:conf/emnlp/RashkinCJVC17}, user comments~\cite{DBLP:conf/kdd/ShuCW0L19}, news environments~\cite{DBLP:conf/acl/ShengCZLWZ22}, or social graphs~\cite{DBLP:conf/cikm/NguyenSNK20}. Sentiment has been recognized as one of the most influential news values in misinformation detection~\cite{dblp:Gullal_news_values}. Zaeem et al.~\cite{DBLP:conf/asunam/ZaeemLB20} observed a statistically significant relationship between negative sentiment and fake news and between positive sentiment and true news. A number of studies have used sentiment-related features, sometimes combined with other content features, for fake news detection. For example, Wang et al.~\cite{DBLP:conf/semeval/WangLW17}, combined sentiment with Twitter metadata features, or Xuan et al.~\cite{DBLP:conf/icdm/XuanX19}, integrated it with content and user features. Although these methods demonstrate the utility of sentiment in misinformation detection, they primarily treat it as a discriminative feature rather than considering its potential misuse as an adversarial manipulation strategy, particularly in the context of LLM-generated content.

\subsection{Adversarial Attacks on Fake News Detectors}
Enhancing the robustness of fake news detectors requires a thorough understanding of their vulnerabilities to different types of adversarial manipulation. Existing studies have examined a range of adversarial attacks including manipulation of social engagement signals~\cite{DBLP:conf/www/WangDC0YS23}, user behavior~\cite{DBLP:conf/kdd/HeAK21}, factual distortion~\cite{DBLP:journals/corr/abs-2107-07970}, subject-object role reversals~\cite{DBLP:conf/icaart/ZhouGBH19}, and data availability constraints~\cite{DBLP:journals/tist/HorneNA20}. In addition, Wu et al.~\cite{DBLP:sheepdog} focused on the style of news publisher and proposed \textit{SheepDog} to evaluate the impact of writing styles against LLM-empowered attacks. However, to the best of our knowledge, the impact of sentiment-based adversarial attacks, particularly those generated by LLMs, 
has not been explored in the context of fake news detection.

\subsection{LLMs for Fake News Detection}
Large language models (LLMs) such as GPT (Generative Pre-trained Transformer)~\cite{chatgpt}, LLaMA (Large Language Model Meta AI)~\cite{DBLP:LLAMA}, Mistral-7B~\cite{DBLP:Mistral} and Qwen~\cite{DBLP:Qwen3} which are trained on large-scale corpora 
have shown impressive emergent abilities on various tasks~\cite{DBLP:journals/tmlr/WeiTBRZBYBZMCHVLDF22}. 
Recent work explored the role of LLMs in fact-checking \cite{DBLP:conf/acl/PanWLLWKN23, DBLP:conf/ijcnlp/ZhangG23, DBLP:conf/cikm/TahmasebiME24,DBLP:conf/ecir/TahmasebiME25} and fake news detection \cite{DBLP:conf/iclr/ChenS24, DBLP:conf/emnlp/PelrineITRGCGR23}. Pan et al.~\cite{DBLP:conf/acl/PanWLLWKN23} introduced ProgramFC, which leverages the in-context learning ability of LLMs to generate reasoning programs to guide the fact checking. They decompose complex claims into simpler sub-tasks, each solved with a shared library of specialized functions. Tahmasebi et al.~\cite{DBLP:conf/cikm/TahmasebiME24} examined the zero-shot capabilities of LLMs for misinformation detection and incorporated a multimodal evidence retrieval component, where retrieved images and texts are re-ranked by large vision–language models (LVLMs) to support fact verification. For fake news detection, Pelrine et al. \cite{DBLP:conf/emnlp/PelrineITRGCGR23} focused on generalization and uncertainty, proposing techniques to leverage LLMs in detecting information veracity under imperfect classification scenarios. However, the impressive strengths of LLMs have also attracted increasing attention towards LLM-generated misinformation \cite{Kreps_McCain_Brundage_2022}. Recent investigations have found that LLMs can act as high-quality misinformation generators \cite{DBLP:conf/acl/HuangMNCJ23, DBLP:conf/emnlp/LucasUYLR023}, and that LLM-generated misinformation is often more deceptive and harder to detect than human-written content and potentially cause more harm \cite{DBLP:conf/iclr/ChenS24, DBLP:journals/aim/ChenS24}.

\section{Robust Fake News Detection using AdSent}
\label{sec:methodology}

\begin{figure*}[t]
  \centering
  \includegraphics[width=0.93\linewidth]{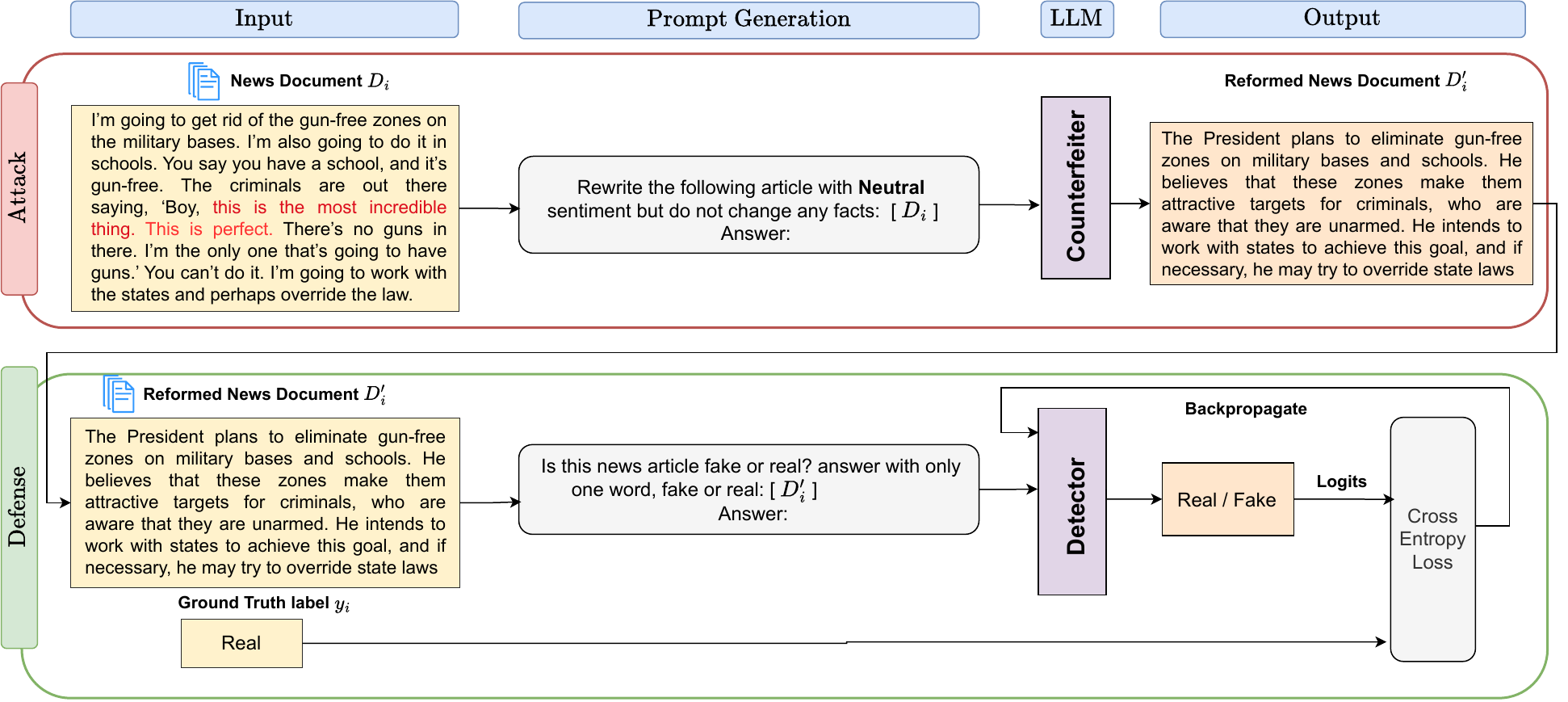}
  \caption{Overview of the proposed AdSent framework for sentiment-robust fake news detection. The framework consists of two main components: an attack module (red box) and a defense module (green box). Given an input document $D_i$, the model predicts whether it is fake or real.}
  \label{pipeline_fig}
\end{figure*}

In this section, we propose \textit{Adsent} pipeline that exploits LLMs for fake news detection. 
We define the task of text-based fake news detection as follows. 
Given a data corpus $\mathbb{C} = \{D_1, D_2, \dots,\ D_m\}$ comprising $m$ news documents with corresponding ground-truth labels~$y \in \{0, 1\}$, the goal is to predict whether an input news document~$D_i$ is real ($y=0$) or fake ($y=1$). 
The proposed AdSent framework, illustrated in Figure~\ref{pipeline_fig} consists of two main components: (1)~a \textbf{sentiment attack component}, which uses LLMs to manipulate the sentiment of a news document while preserving the factual content (Section~\ref{sec:methodology_attacks}), and 
(2)~a \textbf{sentiment-agnostic training} component, 
to fine-tune LLMs to predict the veracity of news documents without being influenced by sentiment features~(Section~\ref{sec:methodology_defense}).

\subsection{Sentiment Attack using LLMs}
\label{sec:methodology_attacks}
In this section, we introduce a strategy to manipulate the news sentiment of each document \(D_i\) using LLMs to obtain sentiment-shifted variants \(\tilde{D}_i\). 
Formally, given a corpus~\(\mathbb{C}=\{D_1,D_2,\dots,D_m\}\) with veracity labels \(y\in\{0,1\}\), our goal 
is to produce three controlled reframed 
version \(\tilde{D}_i^{\text{pos}}, \tilde{D}_i^{\text{neg}}, \tilde{D}_i^{\text{neu}}\) for each document \(D_i\) that vary only in sentiment framing while preserving the factual content and coherence of the original document. %
The general idea is to simulate adversarial scenarios in which an article's sentiment is altered without changing its veracity. 
LLMs are well-suited as counterfeiter for this task, since they can rewrite documents fluently while adjusting their emotional tone in a controlled way. 
To ensure meaningful counterfeiting, the reframing 
must preserve factual details, avoid introducing new claims, and result in coherent, realistic text that resembles how misinformation could be manipulated in practice. To achieve this, we design prompting strategies that explicitly instruct the model to keep all factual information intact while rewriting the document toward a desired sentiment polarity.
The general prompt format is defined as follows:
\begin{mdframed}[backgroundcolor=gray!10,skipabove=5pt,skipbelow=5pt,innertopmargin=2pt,innerbottommargin=2pt,roundcorner=2pt,frametitle={Prompt},frametitlerule=true,frametitleaboveskip=2pt,frametitlebelowskip=2pt]
\noindent \textit{Rewrite the following article with \{positive, negative, neutral\} sentiment but do not change any facts! Also, do not include the prompt in the response and do not summarize or expand the original article!}
\end{mdframed}
\vspace{5pt}
This strategy allows exploring the following research question (RQ):

\textit{RQ1: To what extent can text-based fake news detectors withstand LLM-driven sentiment attacks?}

%
To address RQ1, we define a model-agnostic and reproducible framework for generating and assessing robustness to sentiment-based adversarial attacks. The generated sentiment-shifted variants \(\tilde{D}_i^{\text{pos}}, \tilde{D}_i^{\text{neg}}, \tilde{D}_i^{\text{neu}}\) are supplied to a detector which may be any (large) language model. Detectors are treated as black-box classifiers that process both the original documents \(D_i\) and their reframed variants \(\tilde{D}_i\) under identical preprocessing and inference settings. Robustness is then quantified as the  performance difference between original and manipulated inputs, providing a model-agnostic measure of resilience against sentiment-based adversarial attacks.

\textit{RQ2: Which sentiment makes the fake news more challenging?}

To address this research question, 
we 
suggest a method that analyzes prediction dynamics under sentiment manipulation at the instance level. The central idea is to trace how detector outputs change 
when documents are rephrased with different sentiment polarities. For each news document \(D_i\), we record 
the ground truth~$y_{gt}$, the prediction on the original text~$y_{orig}$, and the prediction under adversarial manipulation~$y_{adv}$ and use the notation scheme $y_{gt}\;y_{orig} \;\rightarrow\; y_{adv}$ to denote potential prediction changes under adversarial attacks. For better comprehension, we denote a real news item ($y=0$) with $R$ and a fake news item ($y=1$) with $F$. For example, $R R \rightarrow F$ indicates a real news item that is correctly classified as 'real' on the original sample but flipped to 'fake' under sentiment manipulation. This formulation results in eight possible prediction scenarios, allowing us to realize how sentiment shifts (positive, negative, neutral) affect classification reliability. We explore potential model biases and identify which sentiment polarities pose the greatest challenge for distinguishing between real and fake news, independent of model architecture. 

\subsection{Sentiment-Agnostic Fake News Detection}
\label{sec:methodology_defense}
In this section, we introduce 
a sentiment-agnostic fake news detection component  
as shown in Figure~\ref{pipeline_fig}. %
The core idea is to create input documents with neutral sentiment using the counterfeiter to train a fake news detection model that solely relies on factual content rather than sentiment cues. 


First, for each article~${D}_i$, we use an LLM as counterfeiter model to create a version~$\tilde{D}_i$ with neutral sentiment  as described in Section~\ref{sec:methodology_attacks}. We choose neutral sentiment as this resembles news reports that aim to objectively report on world events.  
Second, we train a fake news detection model on the corresponding sentiment-altered variants~$\tilde{D}_i$ to mitigate any bias in detector toward sentiment. We formulate the task as a supervised binary classification problem and adopt an LLM such as LLaMA~\cite{DBLP:LLAMA} as detector model, given its extensive world knowledge and demonstrated effectiveness in misinformation detection tasks~\cite{DBLP:conf/cikm/TahmasebiME24, DBLP:conf/pakdd/Srinivasan25, DBLP:conf/www/ZhengZWBLL25}. 
We instruct the LLM-based detection model to predict whether a news article is fake or real using the following prompt. 

\begin{mdframed}[backgroundcolor=gray!10,skipabove=5pt,skipbelow=5pt,innertopmargin=2pt,innerbottommargin=2pt,roundcorner=2pt,frametitle={Prompt},frametitlerule=true,frametitleaboveskip=2pt,frametitlebelowskip=2pt]
\noindent \textit{Is this news article fake or real? Answer only with one word, fake or real : [news article] Answer:}
\end{mdframed}
\vspace{5pt}



\noindent After encoding the input prompt, we extract the logits of the LLM~$\theta$, e.g., LLaMA~\cite{DBLP:LLAMA}, for the tokens \textit{fake} and \textit{real} and use the softmax activation function to create a two-dimensional output vector~$\mathbf{\hat{y}}(\theta) \in [0, 1]^2$ with the corresponding probabilities. We choose the cross-entropy loss between the output vector~$\mathbf{\hat{y}}(\theta)$ to the one-hot encoded ground-truth vector~$\mathbf{y}$ for optimization. 

During inference, we follow the same strategy and first create a neutral version $\tilde{D}_i$ of an article using the counterfeiter, which is used as input for the fake news detection model. Ultimately, we use the generated output, i.e., "real"~($y=0$) or "fake"~($y=1$) to predict the article veracity~$y$. 


\section{Experiments}
\label{sec:experiments}

This section outlines the experimental settings and findings of our study. 
First, we describe the overall experimental setup~(Section~\ref{sec:exp_setup}). 
We then present and analyze the results from two complementary perspectives: the performance of established state-of-the-art models on news articles with manipulated sentiment~(Section~\ref{sec:exp_attack}) and the robustness achieved through sentiment-agnostic finetuning~(Section~\ref{sec:exp_training}). Finally, we conduct a generalization study~(Section~\ref{sec:exp_generalization}), which investigates how well the proposed approach extends to different types of adversarial data as well as to diverse content. 

\subsection{Experimental Setup}
\label{sec:exp_setup}

In this section, we describe the experimental setup adopted to evaluate the effectiveness of our approach. We begin by introducing the datasets for training and evaluation, followed by a description of the baseline models used for comparison. We then provide the implementation details, including hyperparameter settings and training configurations, to ensure reproducibility of our results. Finally, we outline the evaluation metrics employed to assess model performance and robustness.

\subsubsection{Datasets}
We conduct our evaluation on three widely used real-world benchmark datasets
including the \textit{PolitiFact} and \textit{GossipCop} datasets that are part of the FakeNewsNet public benchmark~\cite{DBLP:fakenewsnet}, as well as the \textit{Labeled Unreliable News (LUN)} dataset~\cite{DBLP:lun_dataset}. 
Table~\ref{tab:dataset_stats} summarizes the key statistics of these datasets. PolitiFact and LUN primarily cover political content, with text claims crawled directly from Politifact\footnote{www.politifact.com} websites, whereas GossipCop focuses on celebrity-related gossip news. The LUN dataset originally divides unreliable news into three subcategories: satire, hoax, and propaganda. In our setting, we treat all three as fake and perform binary classification between real and fake news, while considering equal number of samples in both 
classes. To better approximate real-world deployment scenarios, we follow the setup of prior work and use the same splits proposed 
by Wu et al.~\cite{DBLP:sheepdog}. Specifically, 
we apply their temporal data splitting on PolitiFact and GossipCop, where timestamp information is available. The most recent 20\% of news articles (balanced across real and fake instances) are used as the test set, while the earlier 80\% of articles form the training set. For the LUN dataset, we likewise follow Wang et al.~\cite{DBLP:sheepdog} and employ their 
random 80/20 split 
into training and test sets to ensure full comparability with baseline results.
\begin{table}[!t]
\centering
\caption{Statistics of the dataset used in the experiments}
\begin{tabular}{l cc cc cc}
\midrule
\textbf{Dataset} & \multicolumn{2}{c}{\textbf{Train}} & \multicolumn{2}{c}{\textbf{Test}}  & \multicolumn{2}{c}{\textbf{All}}\\
\cmidrule{2-3}\cmidrule{4-5} \cmidrule{6-7}
 & Real & Fake  & Real & Fake  & Real & Fake \\
\midrule
PolitiFact & 180 & 180 & 45 & 45 & 225 & 225\\
\midrule
GossipCop  & 3,166 & 3,166 & 792 & 792 & 3,958 & 3,958\\
\midrule
LUN & 3000 & 3000 & 750 & 750 & 3750 & 3750\\
\bottomrule
\end{tabular}
\label{tab:dataset_stats}
\end{table}

\subsubsection{Baselines}
We compare the performance of AdSent against representative models from three groups. 
The first group, \textbf{fine-tuned LMs~(G1)}, adapts pretrained language models to the fake news detection task and has demonstrated strong effectiveness in handling misinformation scenarios~\cite{DBLP:conf/www/PelrineDR21}. 
Within this group, we include the representative RoBERTa model~\cite{DBLP:roberta}, which has been shown to be one of the most promising approaches in the related literature~\cite{DBLP:sheepdog}. 
This baseline is implemented using the base version of the model, in line with our approach for fake news detection.

The second group, \textbf{fine-tuned adversarial models~(G2)}, comprises approaches that are explicitly fine-tuned on adversarial variants of data in order to improve robustness of misinformation detection. 
As a representative of this group, we consider SheepDog~\cite{DBLP:sheepdog}, which is adversarially fine-tuned with a focus on the style of the news publisher, and propose a style-robust fake news detector that prioritizes content over style in determining news veracity.

The third group, \textbf{zero-shot LLMs~(G3)}, leverages LLMs directly without task-specific fine-tuning. 
These models rely on their extensive pretraining to generalize to fake news detection through carefully designed prompts, enabling evaluation in a zero-shot setting. 
As representatives of this group, we include LLaMA-3.1-8B-Instruct~\cite{DBLP:LLAMA} and Qwen2.5-7B-Instruct~\cite{DBLP:Qwen3}, which reflect two state-of-the-art instruction-tuned LLM families.

\subsubsection{Implementation Details}
We implement AdSent using PyTorch 2.4.1 with CUDA 12.1, and employ the pretrained LLaMA-3.1-8B-Instruct weights from Hugging Face Transformers 4.46.3. 
During training, we use the AdamW optimizer with a learning rate of \(2 \times 10^{-5}\), a batch size of 1 for the PolitiFact dataset and 5 for GossipCop. 
Fine-tuning is performed for five epochs.
For baseline methods, we follow the original architectures and hyperparameter settings proposed by the respective authors~\cite{DBLP:sheepdog, DBLP:roberta}. 
For zero-shot LLM predictions in the sentiment 
manipulation experiment, we apply greedy decoding and perform each experiment once. 
All LLMs are loaded with 8-bit quantization and prompted using the best-performing prompt as described in Section~\ref{sec:methodology}. While we omit the details of prompt exploration for conciseness, we explored multiple prompts during development and used the most effective one across all experiments.
All experiments are conducted on two NVIDIA H100 GPUs with 80 GB memory each. 

\subsubsection{Metrics}
Model performance is evaluated using Accuracy (\%), Precision (\%), Recall (\%), and macro F1-score (\%). To ensure robustness, we report averaged metrics over ten independent runs for all experiments. 
However, for experiments involving LLM-based sentiment manipulation, we report results from a single run since it is very time-consuming to generate long manipulated documents and to manually verify whether the fact are preserved~(see Section~\ref{sec:attack_consistency}).

\subsection{Impact of Sentiment Manipulation}
\label{sec:exp_attack}
In this section, we present the results of representative models from different groups under both the original and adversarial settings (Section~\ref{sec:attack_results_RQ1} and \ref{sec:attack_results_RQ2}) to assess their vulnerability. We further evaluate the consistency of the manipulation process to ensure that the sentiment attack was applied as intended, preserving the factual information, and report the corresponding results (Section~\ref{sec:attack_consistency}).

\subsubsection{Robustness Against Sentiment Attacks}
\label{sec:attack_results_RQ1}
To address the first research question~(RQ1) \textit{``To what extent can text-based fake news detectors withstand
LLM-driven sentiment attacks?''}, we start with our first hypothesis, that real news typically convey a more positive sentiment, whereas fake news tends to be more negative~\cite{DBLP:conf/asunam/ZaeemLB20}. To test this, we apply the sentiment manipulation 
procedure described in Section~\ref{sec:methodology_attacks} to the PolitiFact and Gossipcop test set, altering the sentiment of real news to negative and fake news to positive. 
The objective is therefore to construct a more challenging test set by modifying sentiment while preserving factual content which is guaranteed in fact preservation experiment presented in Section~\ref{sec:attack_consistency}. 
We then evaluate all methods on both original and adversarial test sets with results presented in Table~\ref{tab:reframed-results}.

\begin{table}[t]
\centering
\caption{Macro F1-score (F1) and performance drop of different groups of models on the \textit{Original (Org)} and \textit{Adversarial (Adv)} versions of the Politifact and Gossipcop test sets. The groups include: G1: Fine-tuned LMs, G2: Fine-tuned adversarial models, and G3: Zero-shot LLMs. The adversarial set is constructed by altering the sentiment of real samples to negative and fake samples to positive. 
}
\label{tab:reframed-results}
\begin{tabular}{l l c c c c c}
\toprule
 & & & \multicolumn{2}{c}{\textbf{PolitiFact}} & 
\multicolumn{2}{c}{\textbf{GossipCop}} \\
\cmidrule(lr){4-5} \cmidrule(lr){6-7}
\textbf{Group} & \textbf{Model}
& \textbf{Set}
& \textbf{F1} & \textbf{drop}
& \textbf{F1} & \textbf{drop}
\\
\midrule

\multirow{2}{*}{\textbf{G1}}
& \multirow{2}{*}{RoBERTa}
& Org  & 81.33 & \multirow{2}{*}{\(\text{21.51}\,\downarrow\)} & 74.05 & \multirow{2}{*}{\(\text{16.09}\,\downarrow\)}\\
& & Adv & 59.82 &  & 57.96 &\\

\midrule
\multirow{2}{*}{\textbf{G2}}
& \multirow{2}{*}{SheepDog}
& Org & \textbf{89.92} & \multirow{2}{*}{\(\text{11.48}\,\downarrow\)} & \textbf{77.08}  & \multirow{2}{*}{\(\text{6.44}\,\downarrow\)}\\
& & Adv & \textbf{78.44} & & \textbf{70.64} & \\

\midrule
\multirow{4}{*}{\textbf{G3}}
& \multirow{2}{*}{\makecell[l]{LLaMA-3.1-\\8B-Instruct}}
& Org & 72.66 & \multirow{2}{*}{\textbf{\(\text{4.92}\,\downarrow\)}} & 54.38 &  \multirow{2}{*}{\textbf{\(\text{3.97}\,\downarrow\)}}\\
& & Adv  & 67.74 &  & 50.41 & \\

\cmidrule{2-7}

& \multirow{2}{*}{\makecell[l]{Qwen2.5-\\7B-Instruct}}
& Org & 58.40 & \multirow{2}{*}{\(\text{5.15}\,\downarrow\)} & 46.82 &  \multirow{2}{*}{\(\text{5.47}\,\downarrow\)}\\
& & Adv & 53.25 &  & 41.35 &\\

\bottomrule
\end{tabular}
\end{table}

The results indicate that all state-of-the-art fake news detectors are vulnerable to LLM-driven sentiment attacks. This vulnerability leads to substantial performance degradation, with F1-scores dropping by up to 21.51\% on the Politifact adversarial test set. 
%
On both datasets, While supervised approaches~(G1, G2), in particular SheepDog, provide better results than zero-shot LLMs~(G3), their performance degrades more noticeably on adversarial data. 
This indicates less robustness and more bias in their decisions.  
%
\subsubsection{Impact of Sentiments Variations}
\label{sec:attack_results_RQ2}
We also conduct a more fine-grained analysis of prediction shifts caused by LLM-driven sentiment attacks. To address our second research question~(RQ2)  \textit{``Which sentiment makes fake news more challenging to detect?''}, we perform a deeper analysis using the 
LLaMA-3.1-8B-Instruct model, which has been the best and most robust model in our previous experiment~(see Table~\ref{tab:reframed-results}). 
For this purpose, we evaluate its performance on the original PolitiFact test set as well as on sentiment-altered variants of the same data (positive, negative, and neutral). To quantify the impact of sentiment changes, we measure the number of prediction flips following the procedure described in Section~\ref{sec:methodology_attacks}. The corresponding results are presented in Table~\ref{tab:flip_stats}.

%
The results indicate that, contrary to our initial hypothesis that positive sentiment would be more associated with real news and negative sentiment with fake news, neutral articles are in fact more likely to be classified as real, while non-neutral articles are more likely to be classified as fake. 
A potential explanation is that legitimate news agencies generally aim to remain unbiased in their reporting and thus present events in a neutral manner, whereas fake news often relies on overly emotionalized content (either positive or negative) to persuade readers.
This finding is further supported by the lowest F1 score reported for neutral set and the analysis of prediction flips. 
Specifically, we observe a higher number of cases in which fake news articles that were correctly classified as fake in the original test set are misclassified as real (FF$\rightarrow$R) when their sentiment is altered to neutral. 
This pattern holds consistently across both PolitiFact and GossipCop reporting higher number of FF$\rightarrow$R for the neutral variant than for the positive or negative variants, reinforcing the conclusion that neutral sentiment poses the greatest challenge for fake news detection. 
On the other hand, the number of real samples which are correctly classified as real in the original set and misclassified as fake (RR$\rightarrow$F) when the sentiment is neutralized is lower in neutral set compared to other variants. 
Therefore, we continue with the neutral set as the adversarial setting and conduct further experiments on this variant. 
\begin{table*}[t]
\centering
\caption{
Average macro-F1 scores and prediction flip counts for LLaMA-3.1-8b-Instruct model on different sentiment-altered versions of the Politifact and Gossipcop test sets. Each row corresponds to a specific version of the test set (Original, Positive, Negative, or Neutral sentiment). Arrows indicate how model predictions changed: for example, RR$\rightarrow$F denotes a real sample that was originally predicted as \textbf{real}, but misclassified as \textbf{fake} in the respective test set.
}

\label{tab:flip_stats}

\setlength{\tabcolsep}{8pt}

\begin{tabular}{l l c c c c c c c c c}
\toprule
\textbf{Dataset} & \textbf{Set} & \textbf{Avg F1} 
& \textbf{RR$\rightarrow$R} & \textbf{RR$\rightarrow$F} 
& \textbf{RF$\rightarrow$R} & \textbf{RF$\rightarrow$F} 
& \textbf{FR$\rightarrow$R} & \textbf{FR$\rightarrow$F} 
& \textbf{FF$\rightarrow$R} & \textbf{FF$\rightarrow$F} \\
\midrule

\multirow{4}{*}{Politifact}
& Original & 72.66
 & -- & -- & -- & -- & -- & -- & -- & -- \\
& Positive & 64.16 & 32 & 12  & 0  & 0  & 16  & 7  & 4  & 18 \\
& Negative & 68.87 & 32 & 13 & 0 & 0 & 13 & 10 & 2 & 20 \\
& Neutral  & \textbf{59.33} & 40 & 5  & 0 & 0  & 19  & 4  & 10  & 12  \\
\midrule

\multirow{4}{*}{Gossipcop}
& Original & 54.38 & -- & -- & -- & -- & -- & -- & -- & -- \\
& Positive & 52.29 & 373 & 260  & 18  & 141  & 328  & 197  & 26  & 241 \\
& Negative & 54.29 & 334 & 299 & 27 & 132 & 253 & 272 & 34 & 233 \\
& Neutral  & \textbf{48.70} & 552 & 81  & 78 & 81  & 463  & 62  & 128  & 139  \\
\midrule

\end{tabular}%
\end{table*}

\subsubsection{Consistency Check}
\label{sec:attack_consistency}
\begin{figure}[t]
    \centering
    \includegraphics[width=\columnwidth]{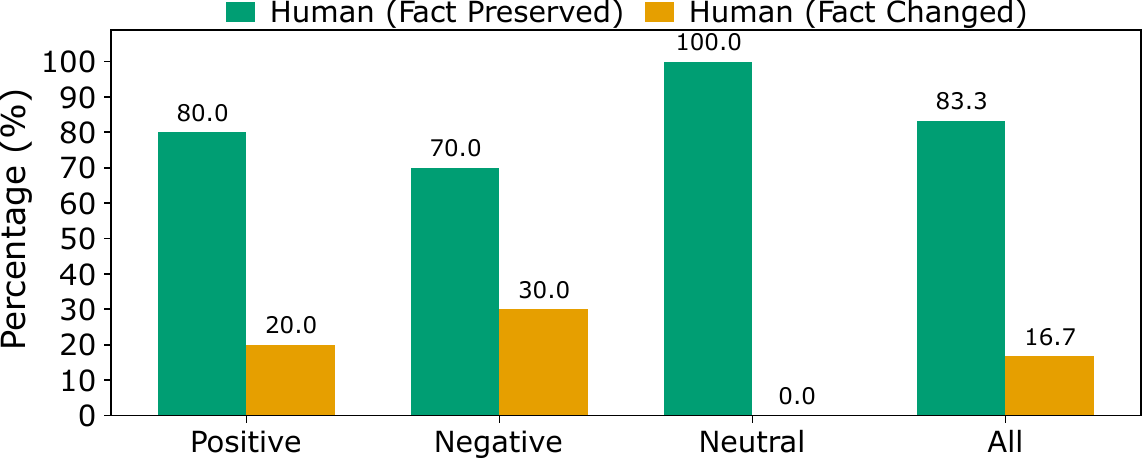} 
    \caption{
    Fact-preservation accuracy (\%) across different annotated variants of the PolitiFact test set, evaluated by humans }
    \label{fig:annotation_bar}
\end{figure}
\begin{figure}[t]
    \centering
    \includegraphics[width=\columnwidth]{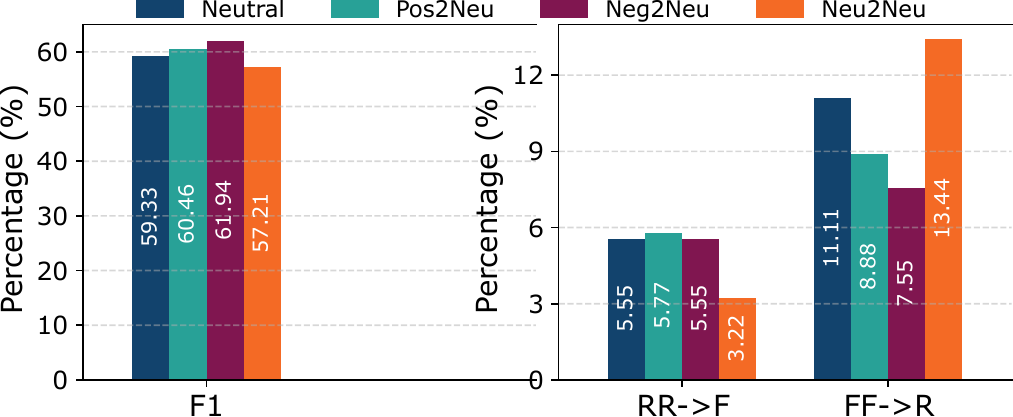}  
    \caption{Macro-F1 (F1) (\%) and percentage of flips in predictions for LLaMA-3.1-8b-Instruct model on different variants of Politifact test set.}
    \label{fig:consistency_bar}
\end{figure}
To ensure that factual content is preserved during the manipulation attack and to provide a more comprehensive assessment of consistency between the original news articles and their manipulated variants, we conduct two complementary experiments.

\textbf{Fact Preservation.} 
This experiment directly evaluates fact preservation under sentiment manipulation using both LLM-based checks and human expert annotations. 

\textit{Human Judge.} We provide a subset of the Politifact test set with manual annotation showing fact preservation between each of the original article compared to its manipulated positive, negative and neutral variants. 
For annotation, an expert in misinformation analysis randomly selected 30 articles, with ten articles from each positive, negative and neutral sets. The expert  manually compared each original article with its sentiment-altered counterpart and assigned a binary label indicating factual preservation: flip = 1 if any factual information changed, and flip = 0 otherwise. 

The results reported in Figure~\ref{fig:annotation_bar} indicate a reasonably high level of consistency between the central factual claims in the original news articles and their sentiment-altered variants. 
Manual annotation shows that factual preservation remains strong with accuracy percentage ranging from 70\% to 100\%, and the neutral sentiment variants which we use for subsequent experiments, achieve the highest accuracy, without a single fact perturbation detected in the annotated subset. 
The lowest score occurs for negative sentiment, where most failures stem from altered quotes whose factual content shifts when sentiment is forced to negative.

\textit{LLM-as-a-Judge.} Once ground-truth labels are established by manual annotation, we investigate if fact preservation can be judged automatically, for scaling to larger datasets. 
To this end, we present the same original/manipulated pairs to an LLM and ask whether factual content is consistent across two versions. We use LLaMA-3.1-8B-Instruct with following prompt and measure agreement between human and LLM annotations using Cohen’s Kappa score.
\begin{mdframed}[backgroundcolor=gray!10,skipabove=5pt,skipbelow=5pt,innertopmargin=2pt,innerbottommargin=2pt,roundcorner=2pt,frametitle={Prompt},frametitlerule=true,frametitleaboveskip=2pt,frametitlebelowskip=2pt]
\noindent \textit{Do the two documents present the same factual information regardless of sentiment? Answer with only one word: yes or no. Document A: [original article] Document B: [manipulated article] Answer: }
\end{mdframed}
\vspace{5pt}
%
The experiment achieves an overall Cohen’s Kappa score of 0.66 for all annotated sets, indicating that LLMs have potential as a judge of factual consistency in detecting whether the core content is preserved under sentiment manipulation. While these results show that LLMs can capture certain factual shifts, they still need further improvement (e.g., via fine-tuning) to serve as a reliable and cost-effective alternative to human annotation.

\textbf{LLM Consistency.} 
This experiment investigates whether LLMs can reliably neutralize sentiment in news articles, confirming the findings reported in Section~\ref{sec:attack_results_RQ2}. To this end, we initially change the sentiment to positive, negative, or neutral using the sentiment attack proposed in Section~\ref{sec:methodology_attacks}. Second, we apply a another manipulation with the counterfeiter model to change the sentiment to neutral. The goal is to verify that LLM-based neutralization is consistent and yields the same conclusions regardless of an article initial sentiment. 

Specifically, we generate the positive, negative, and neutral sentiment variants of all articles in the PolitiFact test set, followed by generating a neutral version based on these variants. This yields three subsets: Positive-to-Neutral (Pos2Neu), Negative-to-Neutral (Neg2Neu), and Neutral-to-Neutral (Neu2Neu), for which we measure prediction flips. The results are shown in Figure~\ref{fig:consistency_bar}.

The results show that second-level sentiment attacks produce results comparable to first-level ones (blue bar in Figure~\ref{fig:consistency_bar}), with the Pos2Neu, Neg2Neu, and Neu2Neu sets achieving nearly the same F1 scores as the original neutral set, differing by no more than ±2\%. In addition, we measure the proportion of real samples correctly predicted as real in the original set but misclassified as fake after manipulation (RR$\rightarrow$F), and fake samples correctly classified as fake in the original set but misclassified as real after manipulation (FF$\rightarrow$R). The percentage of such flips remains approximately consistent across the generated sets compared to the original neutral set, with a maximum deviation of 2.33\%. Notably, the Neu2Neu set shows a 2.3\% increase in FF$\rightarrow$R flips, indicating that second-level neutralization tends to make fake samples more challenging to detect. Overall, these findings confirm that sentiment neutralization directly affects fake news detection while reinforcing the reliability of the observations reported in Section~\ref{sec:attack_results_RQ2}.

\subsection{Impact of Sentiment-Agnostic Training}
\label{sec:exp_training}

\begin{table*}[t]
\centering

\caption{Accuracy (Acc), Precision (Pre), Recall (Rec) and macro F1 (F1) for different groups of models on reformed (neutral sentiment) PolitiFact and GossipCop test set. Best results are highlighted in bold.}
\setlength{\tabcolsep}{4pt}
\begin{tabular}{l l cccccc cccccc}
\toprule
\multirow{3}{*}{\textbf{Group}} & 
\multirow{3}{*}{\textbf{Model}} & 
\multicolumn{6}{c}{\textbf{PolitiFact}} & 
\multicolumn{6}{c}{\textbf{GossipCop}} \\
\cmidrule(lr){3-8} \cmidrule(lr){9-14}
& & Acc & \multicolumn{2}{c}{Real} & \multicolumn{2}{c}{Fake} & F1 & Acc & \multicolumn{2}{c}{Real} & \multicolumn{2}{c}{Fake} &  F1 \\
\cmidrule(lr){4-5}\cmidrule(lr){6-7} \cmidrule(lr){10-11}\cmidrule(lr){12-13}
&&& Pre & Rec & Pre & Rec & & & Pre & Rec & Pre & Rec & \\
\midrule
\multirow{1}{*}{Fined-tuned LMs} 
& DeBERTa & 70.00 & 100.0 & 76.92 & 100.0 & 40.01 & 67.03 & 71.09 & 80.81 & 55.30 & 66.02 & 86.86 & 70.35 \\
\midrule
\multirow{1}{*}{Fine-tuned adversarial} 
& SheepDog & 83.33 & 96.87 & 68.88 & 75.86 & 97.77 & 82.98 & 51.86 & 61.07 & 7.26 & 51.02 & 96.46 & 39.63 \\
\midrule
\multirow{2}{*}{Zero-shot LLMs} 
& LLaMA-3.1 & 62.22 & 57.97 & 88.88 & 76.19 & 35.55 & 59.33 & 52.46 & 51.59 & 79.54 & 55.37 & 25.37 & 48.70\\
& Qwen2.5 & 58.89 & 55.12 & 95.55 & 83.33 & 22.22 & 52.50 & 51.39 & 50.71 & 99.11 & 80.55 & 3.66 & 37.05\\
\midrule
\multirow{1}{*}{Fine-tuned adversarial LLMs}
& AdSent & \textbf{87.77} & 85.41 & 91.11 & 90.47 & 84.44 & \textbf{87.76} & \textbf{79.41} & 73.21 & 88.56 & 81.26 & 72.31 & \textbf{78.33}\\
\bottomrule
\end{tabular}
\label{tab:adsent_results}
\end{table*}

We evaluate AdSent against different groups of baselines on \textit{neutralized} test sets in order to assess veracity detection performance without the influence of sentimental features. 
Results are reported for two datasets in Table~\ref{tab:adsent_results}. 
For each dataset, the fine-tuned LMs are trained on the original training split, SheepDog is trained on the original split augmented with style-altered variants (as described in its official paper), and AdSent is trained on the neutralized variant of the training split.
\begin{figure}[t]
    \centering
    \includegraphics[width=\columnwidth]{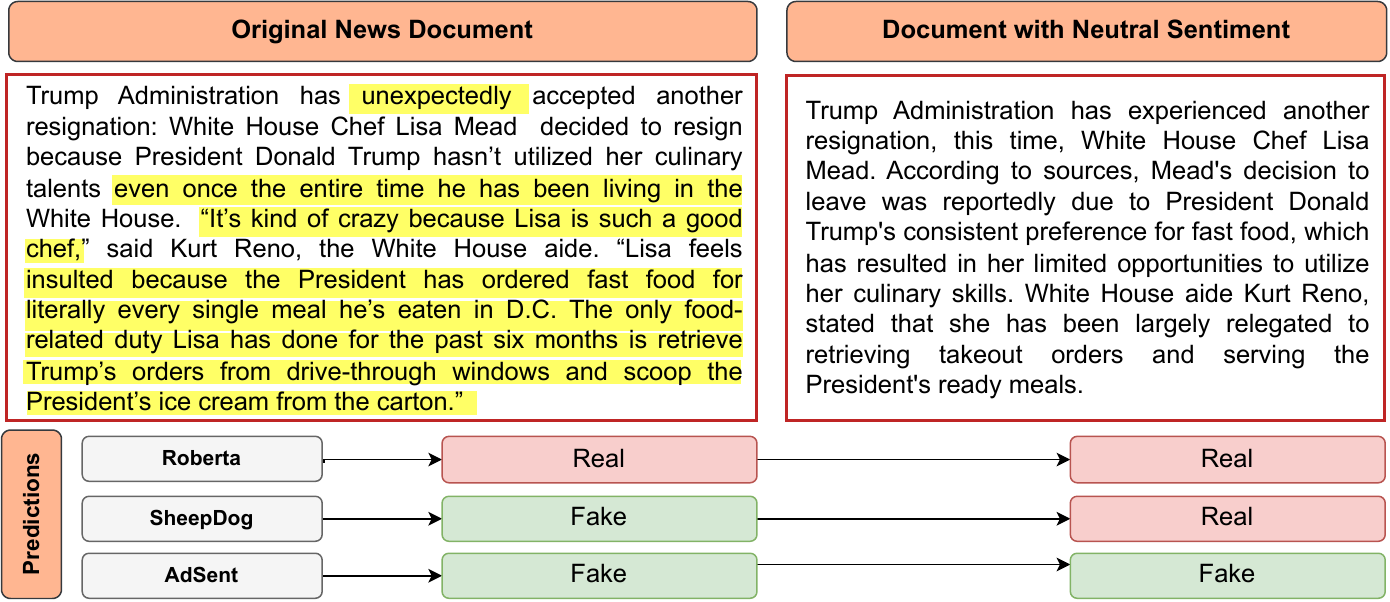}  
    \caption{ Qualitative comparison of three models for adversarial sentiment analysis on Politifact dataset.
    Green text box indicates correct predictions; red text boxes indicates incorrect ones. 
    red borders show fake news document.}
    \label{fig:qualitative_results}
\end{figure}

These results demonstrate that AdSent consistently outperforms all baselines on the neutralized test sets for both PolitiFact and GossipCop achieving the highest macro-F1 of 87.76 and 78.56 respectively. Results also highlight clear differences in how models handle sentiment-neutralized inputs. 
Zero-shot LLMs, such as LLaMA-3.1 and Qwen2.5, show a strong bias toward predicting neutral samples as real class, reflected in very low recall for fake samples. 
This suggests that, without sentiment cues, these models behave more like coarse discriminators of tone rather than robust veracity detectors. 
SheepDog, although adversarially trained, targets style perturbations; once sentiment cues are neutralized, its performance degrades, indicating that robustness to style shifts does not entirely transfer to sentiment confounders. 
In contrast, AdSent trained on neutralized data yields a more balanced behavior, reflecting a sentiment-agnostic decision boundary, as illustrated in Figure~\ref{fig:qualitative_results}.

\subsection{Generalization Study}
\label{sec:exp_generalization}

Generalization is crucial for real-world applications, since misinformation appears in diverse forms and often evolves across domains and platforms. A model that is overly trained to a particular dataset or attack type may struggle when faced with unseen adversarial manipulations or content shifts. To further examine the robustness of our approach, we conduct a generalization study across two complementary dimensions. 
First, we test the model on different types of adversarial data to evaluate whether robustness against sentiment-based attacks extends to other attack strategies such a style of news publisher~(Section~\ref{sec:generalization_adv}). 
Second, we assess its performance on a LUN dataset (Section \ref{sec:generalization_dataset}) to determine whether robustness against sentiment-based attacks can transfer beyond the training distribution and domain. 
\subsubsection{Different Adversarial Data}
\label{sec:generalization_adv}

We further evaluate the generalization ability of our AdSent model on an adversarial variant of the PolitiFact test set proposed in the  SheepDog paper~\cite{DBLP:sheepdog}. 
This benchmark introduces a series of LLM-based style-transfer attacks, where the writing style of news articles is manipulated according to well-known publishers. Specifically, the style of real news is altered to mimic that of \textit{National Enquirer} and \textit{The Sun}, denoted as sets A and B, respectively, while fake news is rewritten in the style of \textit{CNN} and \textit{The New York Times}, denoted as sets C and D, respectively \cite{DBLP:sheepdog}. This procedure results in four distinct adversarial test sets summarized in Table~\ref{tab:adv_style}. 
Notably, the SheepDog method itself fine-tunes a RoBERTa model explicitly for this task, making it a strong reference point for comparison.

The results demonstrate that AdSent consistently outperforms the SheepDog baseline across all adversarial test variants, achieving higher accuracy and macro F1 scores. 
Although SheepDog is specifically designed for this style-transfer attack, AdSent remains effective under the same conditions and even surpasses the specialized baseline. 
Together with the previously observed performance drop of SheepDog on sentiment-based attacks, this indicates that sentiment shifts may have a stronger influence on model predictions than stylistic changes. Consequently, the robustness that AdSent acquires through adversarial sentiment training appears to generalize to style-based attacks as well, allowing the model to maintain strong performance across different types of perturbations.
\begin{table}[t]
\centering
\caption{Performance comparison between AdSent and the SheepDog baseline on adversarial style variants of the Politifact test set introduced by SheepDog~\cite{DBLP:sheepdog}. Metrics reported are Accuracy (Acc), Precision (Pre), Recall (Rec), and macro F1 (F1). The best results are highlighted in bold.}
\setlength{\tabcolsep}{3pt}
\begin{tabular}{l l cccccc}
\toprule
\multirow{2}{*}{\textbf{Model}} & 
\multirow{2}{*}{\textbf{Data}} & \textbf{Acc} & 
\multicolumn{2}{c}{\textbf{Real}} & 
\multicolumn{2}{c}{\textbf{Fake}} & \textbf{F1} \\
\cmidrule(lr){4-5} \cmidrule(lr){6-7}
& &  & Pre & Rec & Pre & Rec & \\
\midrule
\multirow{6}{*}{SheepDog}
& Original & 90.00 & 89.04 & 91.33 & 91.61 & 87.77 & \textbf{89.92} \\ 
& Neutral & 83.33 & 96.87 & 68.88 & 75.86 & 97.77 & 82.98 \\  & Adv-style-A & 81.11 & 82.14 & 80.03 & 81.76 & 82.19 & 80.99\\ & Adv-style-B & 80.00 & 78.04 & 82.03 & 83.29 & 77.97 & 79.89 \\ & Adv-style-C & 82.44 & 85.70 & 79.71 & 80.56 & 85.18 & 82.36 \\ & Adv-style-D & 81.31 & 82.57 & 80.80 & 81.25 & 81.87 & 81.24\\
\midrule

\multirow{6}{*}{AdSent}
& Original & 90.00 & 86.00 & 95.50 & 94.90 & 84.45 & 89.90 \\ 
& Neutral & 87.77 & 85.41 & 91.11 & 90.47 & 84.44 & \textbf{87.76} \\  & Adv-style-A & 85.56 & 92.10 & 77.77 & 80.76 & 93.33 & \textbf{85.47}\\ & Adv-style-B & 84.44 & 89.74 & 77.77 & 80.76 & 93.33 & \textbf{84.38} \\ & Adv-style-C & 87.78 & 92.50 & 82.22 & 84.01 & 93.35 & \textbf{87.74} \\ & Adv-style-D & 86.67 & 90.24 & 82.22 & 83.67 & 91.11 & \textbf{86.64}\\

\bottomrule
\end{tabular}
\label{tab:adv_style}
\end{table}

\subsubsection{Different Content}
\label{sec:generalization_dataset}
\begin{table}[t]
\centering
\caption{Generalization comparison between AdSent and SheepDog baseline trained on Politifact dataset and tested on LUN original and neutralized test set. Metrics are Accuracy (Acc), Precision (Pre), Recall (Rec), and macro F1 (F1). 
}
\setlength{\tabcolsep}{4pt}
\begin{tabular}{l l cccccc}
\toprule
\multirow{2}{*}{\textbf{Model}} & 
\multirow{2}{*}{\textbf{Data}} & \textbf{Acc} & 
\multicolumn{2}{c}{\textbf{Real}} & 
\multicolumn{2}{c}{\textbf{Fake}} & \textbf{F1} \\
\cmidrule(lr){4-5} \cmidrule(lr){6-7}
& &  & Pre & Rec & Pre & Rec & \\
\midrule
\multirow{2}{*}{SheepDog}
& Original & 67.03 & 74.29 & 56.93 & 64.15 & 77.13 & 66.34 \\ 
& Neutral & 61.33 & 69.40 & 45.39 & 58.82 & 77.26 & 59.67 \\  
\midrule
\multirow{2}{*}{AdSent}
& Original & 75.13 & 75.03 & 75.33 & 75.23 & 74.93 & \textbf{75.13} \\ 
& Neutral & 70.67 & 76.36 & 59.86 & 66.99 & 81.46 & \textbf{70.32} \\ 
\bottomrule
\end{tabular}
\label{tab:lun}
\end{table}

To further investigate the generalization capability of our approach, we conduct a cross-dataset study by evaluating AdSent on content outside its training distribution. 
In this setting, both AdSent and 
SheepDog model are trained on the PolitiFact dataset and tested on the LUN dataset. 
Although both datasets focus on political discourse, they differ in the collection process and annotation scheme, providing a suitable setting to examine how well the models can transfer their robustness against sentiment-based perturbations to unseen data. To ensure a comprehensive evaluation, we assess performance not only on the original LUN test set but also on its sentiment-neutralized counterpart. The results are reported in Table~\ref{tab:lun}.

The results indicate that AdSent consistently outperforms the SheepDog baseline across both the original and neutralized versions of the LUN dataset. While both models experience a drop in performance on the neutralized test set, the decrease is smaller for AdSent. This suggests that adversarial sentiment training, combined with the representational power of LLMs, equips AdSent with more transferable features that generalize beyond the training distribution. SheepDog, although adversarially trained, appears to overfit to the characteristics of its training set and therefore struggles more when evaluated on a different dataset.

\section{Conclusions}
\label{sec:conclusion}
In this paper, 
we have systematically examined the robustness of state-of-the-art fake news detectors under sentiment-based adversarial attacks generated by LLMs. 
Through 
a comprehensive experimental analysis, we showed that even the strongest text-based detectors experience substantial performance degradation when sentiment is manipulated, revealing a critical weakness in current approaches. 
Our analysis further highlighted that neutral-toned content is particularly challenging, exposing strong biases in existing models toward classifying such news as real. 
To address these issues, we introduced AdSent, a sentiment-robust detection framework that ensures consistent veracity predictions across sentiment variants of the same article. Extensive experiments on three benchmark datasets demonstrated that AdSent significantly improves both accuracy and robustness compared to competitive baselines, while also generalizing effectively to unseen datasets and adversarial scenarios.

In future work, we will focus on extending AdSent to multimodal misinformation detection, where textual and visual signals jointly shape how news content is interpreted. In particular, we plan to examine how sentiment is conveyed in the visual modality, such as emotional expressions in images, and analyze its interaction with textual sentiment in influencing model predictions. Moreover, we will explore adversarial perturbations targeting other news values, such as proximity, novelty, or prominence, which are key journalistic dimensions shaping the perceived credibility and impact of news, and evaluate fake news detectors under these conditions.
\bibliographystyle{ACM-Reference-Format}
\bibliography{main}


\end{document}